\documentclass[conference]{IEEEtran}
\IEEEoverridecommandlockouts
\usepackage{cite}
\usepackage{amsmath,amssymb,amsfonts}
\usepackage{algorithmic}
\usepackage{algorithm}
\usepackage{graphicx}
\usepackage{textcomp}
\usepackage{xcolor}
\usepackage{booktabs}

\usepackage{tikz}
\usetikzlibrary{decorations.pathreplacing, calc}

\def\BibTeX{{\rm B\kern-.05em{\sc i\kern-.025em b}\kern-.08em
    T\kern-.1667em\lower.7ex\hbox{E}\kern-.125emX}}

\begin{document}

\author{
 
\author{\IEEEauthorblockN{Jingang Jin, Yue Ma, Zhenhang Zhang and Qinru Qiu}
\IEEEauthorblockA{\textit{Department of Electrical Engineering and Computer Science}\\
\textit{Syracuse University, Syracuse, New York, USA}\\
\{jjin24, yma183, zzhan281, qiqiu\}@syr.edu}
}
 
}


\title{An FPGA-Based Spiking Neural Network for Spatiotemporal Pattern Learning and Classification}

\maketitle

\begin{abstract}
Spiking Neural Networks (SNNs) are promising for energy-efficient spatiotemporal pattern recognition at the edge, but efficient deployment and on-device adaptation require specialized hardware.
In this work, we present an FPGA implementation of multilayer SNNs for spatiotemporal pattern classification,  in which both inference and training take place on the same hardware. Our design adopts a temporal-forward-only training algorithm that avoids the memory and latency overhead associated with traditional Backpropagation Through Time (BPTT). 
A mixed-precision scheme balances training stability with resource efficiency, and architectural optimization maximizes hardware and memory utilization. We further identify and resolve hazards that arise when inference and learning share the same memory, including port conflicts between the forward and backward access patterns and read-after-write dependencies in floating-point potential accumulation. 
Experimental evaluations on MNIST and five multivariate time-series classification datasets demonstrate that, with fully on-chip training, our SNN model achieves accuracy comparable to SNN or LSTM trained with BPTT. Compared to CPU and GPU based implementations, our FPGA design delivers 2.5× to 6× higher throughput while consuming an order of magnitude less energy per sample.
\end{abstract}

\begin{IEEEkeywords}
Neuromorphic Computing, Spiking Neural Network, In-hardware learning, FPGA
\end{IEEEkeywords}

\section{Introduction}
Driven by the emergence of cyber-physical systems in recent years, a growing number of neural network models have been integrated into edge devices, such as smart vehicles and wearable sensors. 
Because user behavior, environmental conditions, and even sensor characteristics vary over time and across systems, it is highly desirable for neural networks to adapt to their specific hardware and application environment through in-hardware learning after deployment, in order to maintain performance.

There is also increasing demand for edge devices to process temporal sensor readings not as static, frame-based input, but as a stream with spatiotemporal structure. Compared to buffering an entire sequence and classifying all temporal samples at once, a state-space model (e.g., a recurrent neural network (RNN)) that processes input sample-by-sample incurs much lower computational cost and latency, making it more suitable for edge applications. However, a key challenge is that RNNs are trained using Backpropagation Through Time (BPTT), which unrolls the network over time, so its memory requirement grows linearly with sequence length. Moreover, training cannot begin until the entire sequence has been received, which again causes high latency and a bursty workload. These properties make it difficult to train RNNs on edge devices.

Spiking neural networks (SNNs)\cite{b1}\cite{b2}  present a promising alternative to conventional artificial neural networks. They are extremely energy efficient due to their event-driven dynamics and sparse computation. They are also well suited for temporal pattern recognition, since each neuron maintains a hidden state, making the network inherently an RNN. Finally, recent research on three-factor Hebbian learning \cite{b30} has made it possible to train such networks for spatiotemporal classification without unrolling the network in time \cite{b31}. Hence, unlike BPTT training, where memory requirements grow with the length of the input sequence, memory requirements under these algorithms remain constant throughout training.

Given that their computational patterns differ significantly from those of conventional ANNs, SNNs are not well-suited to traditional platforms like CPUs and GPUs, whose architectures are not optimized for sparse, event-driven, fine-grained parallelism. To leverage the full advantages of SNNs, researchers have proposed several dedicated neuromorphic processors, notably IBM TrueNorth\cite{b3} and Intel Loihi\cite{b4}. Although highly optimized, these neuromorphic
chips often impose limitations on data precision, neuron
models, and supported learning rules. As an emerging research area, SNN structure and training algorithms are evolving rapidly. These new dataflows and computing models require specialized hardware circuit, which current neuromorphic processors cannot efficiently support without approximation and simplification. 

Field-Programmable Gate Arrays (FPGAs) occupy a strategic middle ground, offering a compelling balance of flexibility and efficiency. 
They also provide massive parallelism similar to GPUs, but at a more fine-grained level. Researchers have proposed numerous designs for SNN acceleration on FPGAs \cite{b5}\cite{b6}\cite{b7}\cite{b8}, which exhibit significantly lower energy consumption than their CPU or GPU counterparts while offering greater flexibility than ASIC-based neuromorphic processors. However, existing work primarily focuses on accelerating the inference of pre-trained SNN models, largely overlooking online learning scenarios that are prevalent in edge applications.

When online learning is taken into consideration, SNN implementation complexity increases substantially. Numerous variables, including parameters and neuron potentials, are shared between the forward and backward propagation stages and accessed in diverse patterns, introducing both data and resource (e.g., memory port) dependencies. These dependencies hinder efficient hardware utilization on FPGAs. 

In this work, we present an FPGA implementation of multilayer SNNs with on-chip learning capability. Our design integrates a stateful synapse filter structure within the spiking neuron model, a temporal forward-only training algorithm that eliminates the need for BPTT, and a conservative mixed-precision scheme to achieve high energy efficiency and processing throughput. We further introduce optimization strategies, such as loop unrolling, memory partition and read/write scheduling, for computation and memory access concurrency to fully exploit FPGA hardware resources. Memory hazards that are unique to systems hosting both forward and backward propagation are also identified and resolved. 

Experimental evaluations on MNIST and five multivariate time-series classification datasets demonstrate that our implementation achieves accuracy comparable to SNN or ANN trained using BPTT. Compared to CPU- and GPU-based  implementations, our FPGA acceleration delivers 2.5× to 6× higher throughput at roughly one-tenth the energy per sample.

The contributions of this work are summarized as follows:
\begin{itemize}
    \item We implement multilayer SNNs on FPGAs with in-hardware learning capability, performing both training and inference entirely on-chip for spatiotemporal stream classification.
    \item We propose a resource-aware loop unrolling strategy guided by the empirical observation that both resource consumption and execution cycles scale linearly with the unroll factor, and a coarse-grained pipelining strategy that overlaps weight updates with backward passes.
    \item We propose a BRAM partitioning and interleaving scheme to resolve memory access conflicts between forward and backward passes, and use buffer insertion to resolve Read-After-Write (RAW) hazards.
    \item We evaluate our design on both static image (MNIST) and multivariate time-series classification tasks, demonstrating competitive accuracy against prior on-chip learning implementations and significant efficiency gains over CPU and GPU platforms.
\end{itemize}

\section{Background and Related Works}
\subsection{Neuron Model}
The Leaky-Integrate-and-Fire (LIF)\cite{b12} model is a widely utilized spiking neuron model. Its membrane potential performs a leaky integration of the synaptic input:
\begin{equation}
V[t] = \lambda V[t-1] + \sum_{i} w_i X_i[t] - V_{th} O[t-1]
\end{equation}
where $V[t]$ represents the membrane potential of the neuron at time $t$, also considered as the state of the neuron, and $w_i$ is the synaptic weight corresponding to the $i$-th input, $X_i[t]$. The neuron's output, $O[t]$, is a binary value determined by the Heaviside activation function $H(\cdot)$ and the threshold $V_{th}$, as detailed by the following expression:
\begin{equation}
O[t] = H(V[t] - V_{th}) \begin{cases} 0, & V[t] \leq V_{th} \\ 1, & \text{otherwise} \end{cases}
\end{equation}
 
\begin{figure}[t]
  \centering
  \includegraphics[width=0.6\linewidth]{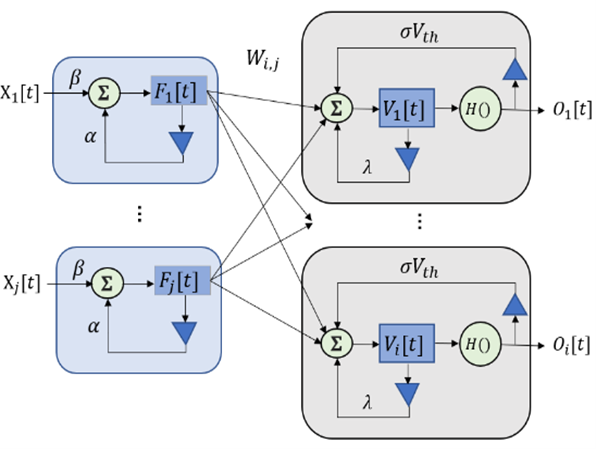}
  \caption{Neuron model of the multilayer SNN}
  \label{fig:neuron_structure}
\end{figure}
 
Recent neuroscience research\cite{b13} suggests that individual neurons' dendritic trees can perform localized computation, where a single neuron is functionally equivalent to an RNN capable of complex temporal responses. An enhanced neuron model incorporating the Post-Synaptic Potential (PSP) was proposed in \cite{b14}, where PSP is defined as the output of a synapse filter governed by a generic IIR (Infinite Impulse Response) kernel with multiple forward taps and backward taps. \cite{b15} and \cite{b16} further demonstrated that competitive performance is achievable even with a first-order filter employing only one forward tap and one backward tap.

In this work, we use a first-order IIR filter for PSP. For the $i$-th input $X_i[t]$ at time $t$, the PSP of the synapse is given by the following equation:
\begin{equation}
F_i[t] = \alpha F_i[t-1] + \beta X_i[t].
\end{equation}
 
The coefficients $\alpha$ and $\beta$ control the filter kernel and are shared by all neurons in our design. With this filter structure incorporated, the membrane potential $V_j[t]$ of the $j$-th neuron at time $t$ is updated according to the following equation:
\begin{equation}
  V_j[t] = \lambda V_j[t-1] + \sum_{i} w_{i,j} F_i[t] - \zeta V_{th} O_j[t-1].
\end{equation}

Here, $\zeta$ is a reset coefficient that controls the strength of the post-spike reset and is set to 0.5 in all experiments. Fig. \ref{fig:neuron_structure} illustrates the neuron structure in one layer. 
 
\subsection{Online Training of SNNs}
\label{sec:solsa}
Online learning\cite{b19} and continual learning\cite{b17} are active research areas, but the common method for training SNNs poses a challenge. SNNs are often treated as a special class of RNN due to their internal memory and temporal filtering, making BPTT the standard training practice. However, BPTT’s requirement to unroll the network across the entire input sequence results in computational and memory demands incompatible with resource-constrained edge devices. To overcome this limitation, learning algorithms such as SOLSA\cite{b15}, OTTT\cite{b18}, and e-prop learning [31] have been proposed for non-BPTT, gradient-based optimization of SNNs. These algorithms are based on similar principles that accumulate temporal gradients in a forward-only manner to avoid the need for unrolling the computation graph. 
 
In this work, we adopt the SOLSA\cite{b15} learning algorithm. For each weight coefficient $w_{i,j}$ from pre-synaptic neuron $i$ to post-synaptic neuron $j$, SOLSA maintains a partial gradient $g_{i,j}[t]$ using the following equation: 
\begin{equation}
g_{i,j}[t] = \sum_{t^{'}= 0}^{t} {\mu_{j}[t^{'}] \times \varepsilon_{i,j}[t^{'}]}
\end{equation}

where $\mu_{j}[t]$ is the spatial (across-layer) gradient of the loss with respect to neuron $j$'s membrane potential, and $\varepsilon_{i,j}[t]$ is an accumulated trace on the $ij$-th post-synaptic potential (the $ij$-th input after the dendritic filter) that carries the temporal history. The trace is updated incrementally as $\varepsilon_{i,j}[t] = \big(\lambda - \zeta V_{th}\,H'(V_j[t])\big)\,\varepsilon_{i,j}[t-1] + F_i[t]$, where $H'(\cdot)$ is the surrogate derivative of the non-differentiable Heaviside firing function, taken as a Gaussian centered at $V_{th}$ with width $\sigma=0.5$. The spatial gradient is obtained by error backpropagation within a single time step, without unrolling over time: for an output neuron, $\mu_j[t] = (\partial \mathcal{L}[t]/\partial O_j[t])\,H'(V_j[t])$; for a hidden neuron, it is propagated from the neurons $k$ in the following layer, $\mu_j[t] = \big(\sum_{k} \mu_k[t]\, w_{j,k}\big)\,H'(V_j[t])$. Since $\mu_j[t]$ depends only on current-time-step quantities, it is evaluated in the same forward-only pass as $\varepsilon_{i,j}[t]$, which alone carries the temporal history.

 
\subsection{FPGA implementations of SNNs}
Driven by the efficiency of SNNs and the flexibility of FPGAs, many works have accelerated SNN inference on FPGA hardware. SyncNN [6] restructures event-driven SNNs into a synchronous dataflow for regular, hardware-friendly computation. FireFly v2 [11], building on its predecessor [8], exploits parallelism across input channels, output channels, pixels, and timesteps, and adds native hardware support for non-spike operations. The work in [7] co-designs the input encoding, neuron model, and architecture to achieve state-of-the-art accuracy and latency. While effective, these works deploy pre-trained SNNs on FPGAs without in-hardware learning.
 
Only a limited number of studies have explored on-chip SNN learning on FPGAs, and most of them rely on biologically inspired but relatively simple learning rules. CORDIC-SNN [22] implements STDP learning with Izhikevich neurons using CORDIC-based arithmetic to replace expensive multipliers with shift-and-add operations, but is limited to a two-layer topology with a single output neuron. TripleBrain [23] combines self-organizing maps with reinforcement-modulated STDP in a fully event-driven hardware core, achieving competitive accuracy on MNIST at low resource cost, yet remains restricted to a single trainable layer. These STDP-based approaches, while hardware-friendly, are inherently unsupervised or require reward signals, hence have limited applications in multilayer supervised learning.

Beyond STDP, the work in [26] takes a fundamentally different approach by employing evolutionary algorithms as a black-box training method for Izhikevich-neuron-based SNNs on a Zynq SoC, enabling continual in-field adaptation without gradient computation. However, evolutionary methods typically suffer from slow convergence and scale poorly with network size, making them impractical for larger architectures. 
HaSiST [27] implements backpropagation on FPGA using a hard-sigmoid surrogate derivative, achieving high accuracy on MNIST, but the training procedure operates on a surrogate ANN that is subsequently converted to an SNN, rather than training the SNN directly. We instead train multilayer SNNs directly on FPGA with no ANN-to-SNN conversion.

\section{Methodology}
\subsection{Design Rationale and Architecture Overview}
Our proposed implementation comprises two core components: a datapath implemented using on-chip memory and programmable logic, and a workflow controller running on the embedded ARM processor. The controller streams in samples from the input time series and collects the classification results. It also provides labels for loss calculation during training. The datapath consists of per-layer forward propagation, backward propagation, and update modules, plus a loss calculation module. We adopted a High Level Synthesis (HLS) based design flow in this work.

SOLSA learning performs both forward and backward propagation once per input sample, doubling the volume of memory accesses compared to an inference only system. On the other hand, since it does not require unrolling the network or storing the history of neuron activities, it has much lower storage requirements than conventional learning algorithms. This allows us to train networks of decent size while constraining the neuron states and weight parameters to fit within the FPGA's on-chip Block RAM (BRAM), rather than relying on off-chip DRAM.

\subsection{Mixed Data Precision}
Prior FPGA acceleration of SNN inference exploited aggressive quantization, reducing weights to 4-bit fixed-point [6], ternary [28], or integer-only [29] representations. However, in-hardware learning imposes tighter precision requirements, as quantization errors accumulate through gradient computation and weight updates, degrading training convergence. We therefore adopt a conservative mixed-precision scheme: FP32 for neuron potential ($V_j[t]$) and FP16 for weight storage ($w_{ij}$). This decision is to balance the precision and memory size. Since a fully connected layer with N inputs and M neurons requires N×M weights but only M potentials, weights dominate the memory footprint. Applying FP16 to them yields a substantial reduction in both storage and resource utilization, while FP32 for the comparatively few potentials ensures stable training. For the similar reason, we adopt FP32 for $F_i$, $\varepsilon_{i,j}$, $\mu_{j}$ and $g_{i,j}$ introduced in Section~\ref{sec:solsa}.

The binary nature of spikes enables additional simplification: using only 1 bit for the spike reduces the multiply-accumulate (MAC) operation on spike input to simple summations, and the spike output of the activation function can be represented by the negation of the sign bit of the difference between membrane potential and threshold.



\subsection{Loop Unrolling}
SNN computation involves extensive vector and matrix operations, which are initially implemented as nested loops in HLS. While sequential execution of these loops induces high latency, 
unrolling all loops for maximum parallelism is infeasible due to limited hardware resources. The challenge is to determine the optimal unroll factor for each loop under a constrained resource budget. To address this, we profile each loop by synthesizing it at different unroll factors and recording the resulting execution time (measured in clock cycles) and resource utilization. For nested loop, only inner loop is unrolled due to hardware resource limitation. The results demonstrate a linear relationship between the unroll factor and both resource consumption and execution time. Based on this profiling, two linear estimation models, $C_l(U_l)$ and $R_l(U_l)$, are trained for each loop $l$ to predict its execution time and resource usage, respectively, under unroll factor $U_l$ . 

Guided by these models, we search for unroll factors using the greedy strategy depicted in Algorithm~1. It repeatedly increments the unroll factor of the loop with the highest estimated execution time, until the performance target is reached or no loop can be unrolled further without exceeding the hardware constraint. This directs hardware resources toward the current bottleneck at each step. 


\begin{algorithm}
\caption{Resource-Aware Loop Unrolling}
\label{alg:unroll}
\begin{algorithmic}[1]
\REQUIRE Loop set $L$, resource budget $R_{\max}$, target latency $C_{\text{target}}$, Linear coefficients for $C_{l}(U_{l})$ and $R_{l}(U_{l})$
\STATE In Unroll Factors $U$, Initialize $U_{l} \gets 1$ for each loop $l \in L$
 \STATE Sort $L$ by $C_{l}(U_{l})$ in descending order
\WHILE{\textbf{true}}
    \STATE $l^{*} \gets L[1]$
    \STATE $U_{l^{*}} \gets U_{l^{*}} + 1$
    \STATE Estimate $C_{l^{*}}(U_{l^{*}})$ and $R_{\text{total}}$ using linear model
    \IF{$\sum_{l \in L} C_{l}(U_{l}) \leq C_{\text{target}}$}
        \STATE Target Achieved
        \STATE \textbf{break}
    \ENDIF
    \IF{$R_{\text{total}} > R_{\max}$}
        \STATE $U_{l^{*}} \gets U_{l^{*}} - 1$
        \STATE Remove $l^{*}$ from L
    \ELSE
     \STATE Insert $C_{l^{*}}(U_{l^{*}})$ back to the sorted list
    \ENDIF
    
\ENDWHILE
\RETURN $U$
\end{algorithmic}
\end{algorithm}

\subsection{Coarse-grained Pipelining}
\begin{figure}[t]
\centering
\begin{tikzpicture}[
    x=1cm, y=1cm,
    block/.style={draw, rounded corners=2pt, inner xsep=4pt, inner ysep=2.5pt, font=\scriptsize, anchor=center},
    fwd/.style={block, fill=blue!20, draw=blue!40},
    bwd/.style={block, fill=red!15, draw=red!35},
    upd/.style={block, fill=violet!15, draw=violet!35},
    lossb/.style={block, fill=gray!15, draw=gray!40},
    io/.style={block, fill=green!10, draw=green!40},
    dep/.style={-stealth, thick, draw=black!60},
    branch/.style={-stealth, thick, draw=black!60},
    ctrl/.style={-stealth, thick, dashed, draw=teal!70},
    acc/.style={font=\tiny, inner sep=1pt},
    mem/.style={draw, fill=orange!15, draw=orange!60, rounded corners=2pt, align=center, font=\scriptsize, inner ysep=3pt, minimum width=5.2cm},
    read/.style={-stealth, thick, densely dotted},
]
\node[io]  (input) at (0, 0) {Input};
\node[fwd] (l1f) at (0, -0.8) {L1 forward};
\node[fwd] (l2f) at (0, -1.6) {L2 forward};
\node[fwd] (l3f) at (0, -2.4) {L3 forward};
\node[lossb] (loss) at (2.2, -2.4) {Loss};
\node[io]  (output) at (2.2, -1.5) {Output};
\node[bwd] (l1b) at (4.4, -0.8) {L1 backward};
\node[bwd] (l2b) at (4.4, -1.6) {L2 backward};
\node[bwd] (l3b) at (4.4, -2.4) {L3 backward};
\node[upd] (l1u) at (6.9,  0.0) {L1 update};
\node[upd] (l2u) at (6.9, -0.8) {L2 update};
\node[upd] (l3u) at (6.9, -1.6) {L3 update};
\draw[dep] (input) -- (l1f);
\draw[dep] (l1f) -- (l2f);
\draw[dep] (l2f) -- (l3f);
\draw[dep] (l3f) -- (loss);
\draw[dep] (loss) -- (l3b);
\draw[dep] (l3b) -- (l2b);
\draw[dep] (l2b) -- (l1b);
\draw[branch] (4.4,-2.0) -| (l3u.south);   
\draw[branch] (4.4,-1.2) -| (l2u.south);   
\draw[dep] (l1b.north) |- (l1u.west);
\draw[ctrl] ([xshift=0.55cm]l3f.east) |- (output.west);
\node[acc, text=teal!70, anchor=south] at ([yshift=1pt]output.north) {inference: skip bwd/upd};
\node[mem] (wmem) at (2.5, -3.5) {Shared weight matrix $W$ (on-chip BRAM)};
\draw[read, blue!65] (l3f.south |- wmem.north) -- (l3f.south);
\node[acc, text=blue!65, anchor=east] at (-0.1, -2.95) {read row-major ($W$)};
\draw[read, red!65] (l3b.south |- wmem.north) -- (l3b.south);
\node[acc, text=red!65, anchor=west] at (4.5, -2.95) {read col-major ($W^{\top}$)};
\end{tikzpicture}
\caption{Architecture and data flow of the multilayer SNN FPGA implementation. 
The forward and backward passes read the \emph{same} on-chip weight matrix in conflicting orders---row-major ($W$) versus column-major ($W^{\top}$)---motivating the partitioning and interleaving scheme.}
\label{fig:arch}
\end{figure}
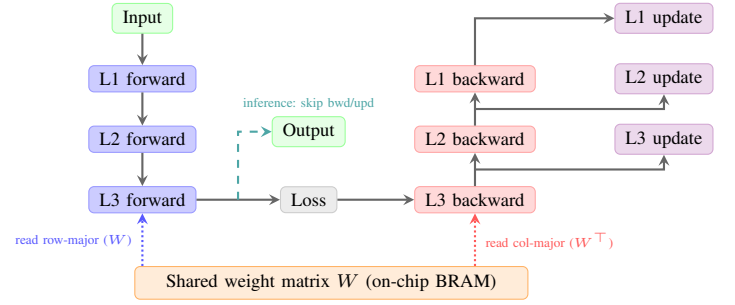

In addition to the loop unrolling, we also consider coarse-grained datapath pipelining. Fig.~\ref{fig:arch} shows the data dependencies among all modules. In the training path, the forward modules execute sequentially ($L_1 \rightarrow L_2 \rightarrow L_3$), as each layer consumes the output of the previous one, and after loss calculation the backward modules execute in reverse order ($L_3 \rightarrow L_2 \rightarrow L_1$), since each layer's gradient depends on the gradient propagated from the layer above. Each layer's weight update, however, depends only on its own backward module, hence can be executed in parallel with the backward module of the next layer as shown in Fig.~\ref{fig:arch}. 

The module-level overlap is fixed at synthesis time by the HLS dataflow structure. Unlike an inference-only system, where the entire forward propagation path can be fully pipelined, the backward path's data dependencies limit the opportunity for coarse-grained pipelining in our design. However, hardware resources can still be fully utilized through loop unrolling.

\subsection{Memory Partitioning}

There are some unique hazards associated to running both inference and learning on the same
hardware. First, the forward and backward passes access the same weight matrix, but in conflicting orders: the forward path reads it row-major, while the backward path reads it column-major. This conflict is exacerbated by loop unrolling. Each BRAM block has only two I/O ports, so with an unroll factor of k, the module must fetch k weights per cycle; if those k elements fall within the same block, the accesses are serialized, negating the benefit of unrolling. Because the two passes require opposing access patterns, no single memory layout can eliminate conflicts for both simultaneously. A layout optimized for the forward pass reintroduces conflicts for the backward pass, and vice versa. An inference-only design, which performs only the forward pass, never encounters this conflict.

\begin{figure}[t]
  \centering
  \includegraphics[width=0.9\linewidth]{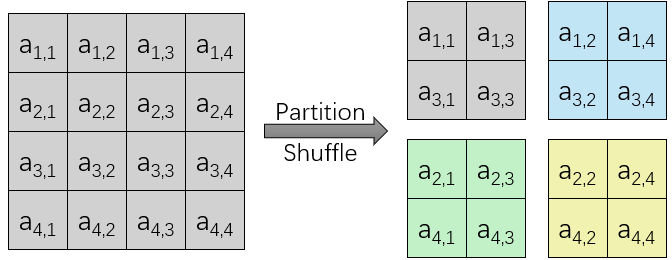}
  \caption{Weight matrix partition and shuffle}
  \label{fig:memory_partition}
\end{figure}

We resolve this hazard with memory partitioning and interleaving. The weight
matrix is divided into $k \times k$ tiles, each mapped to a separate BRAM
block with independent I/O ports; element $w_{i,j}$ is assigned to tile
$[\,i \bmod k,\ j \bmod k\,]$, as illustrated in
Fig.~\ref{fig:memory_partition}. Any $k$ consecutive elements along either
a row or a column are thus distributed across $k$ distinct blocks, so both
the forward and backward passes fetch $k$ elements in a single cycle
without port conflicts.

\subsection{Buffer Insertion}
The second hazard unique to supporting on-chip learning is the Read-After-Write (RAW)
dependency in the accumulating floating point values, such as $V_j = \sum_{i} w_{i,j} F_i[t]$. Learning on-chip requires high precision membrane potentials. 

\begin{figure}[t]
  \centering
  \includegraphics[width=\linewidth]{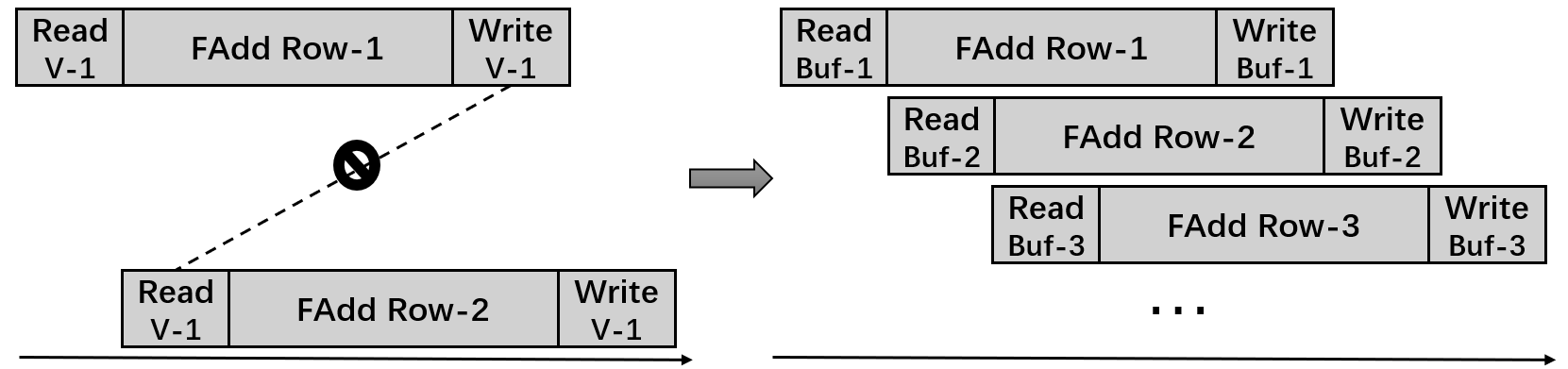}
  \caption{Resolution of RAW hazard with buffer}
  \label{fig:buffer}
\end{figure}

Single precision (FP32) floating point adder on Xilinx FPGA typically has a pipeline latency of 3–5 cycles. As shown in Fig.~\ref{fig:buffer} (a), every iteration updates the same accumulator, iteration $k$ must read the
value written by iteration $k{-}1$; and because the floating-point adder
has a multi-cycle latency $L$, that value is not ready in the next $L$ cycles. Such RAW hazard will not occur in inference only system where fixed-point precision is typically used.

HLS automatically resolves such RAW hazards by raising the pipeline's initiation interval (II) of the pipeline to $L$. However, this serializes the accumulation and negates the benefit of pipelining.
We remove this dependency using an intermediate buffer that splits the
accumulator into $P$ partial sums: iteration $k$ accumulates into line
$(k \bmod P)$, so each line is updated only once every $P$ iterations
(Fig.~\ref{fig:buffer} (b)). The dependency distance per line is then $P$
cycles at II$=1$, and choosing $P \ge L$ sustains II$=1$ regardless of the
reduction length. Since the measured latency of the floating point adder is about four cycles, four buffer
lines suffice. A final step sums the partial sums into the potential at
negligible cost.

\section{Experimental Results}
The experiments were conducted using the AMD ZCU102 Evaluation Board. This platform features a heterogeneous System-on-Chip that integrates a quad-core ARM Cortex-A53 processor with FPGA programmable logic. We developed the hardware implementation using the AMD Vivado Design Suite and Vitis Core Development Kit (version 2024.2).

\subsection{Datasets and Hyper Parameters}
We evaluated our proposed method on six classification datasets spanning both static image and multivariate time-series. Table \ref{tab:datasets} summarizes the key information of these datasets.
MNIST is a widely used benchmark of handwritten digit images ($28\times28$ grayscale), included to enable direct comparison with prior FPGA-based on-chip learning implementations. The remaining five datasets are multivariate time-series from the UEA archive [20], spanning motion sensing, biomedical signal processing, and physiological monitoring applications representative of edge computing scenarios. 
For SOLSA learning, we set both $\alpha$ and $\beta$ in the PSP filter to be 0.9 and $\lambda$ to be 0.5. The learning rate is set to be 0.0001.

\begin{table}[t]
\centering
\caption{Dataset Information}
\label{tab:datasets}
\begin{tabular}{lcccc}
\hline
\textbf{Dataset} & \textbf{Input Dim} & \textbf{Length} & \textbf{Classes} & \textbf{Type} \\
\hline
MNIST            & 784  & 1    & 10   & Static   \\
BasicMotions[20]     & 6  & 100  & 4  & Temporal \\
Epilepsy[24]         & 3  & 207  & 4  & Temporal \\
FingerMovements[25]  & 28  & 50  & 2  & Temporal \\
Racket Sports[20]    & 6    & 30   & 4    & Temporal \\
Japanese Vowels[21]  & 12   & 29   & 9    & Temporal \\
\hline
\end{tabular}
\end{table}

\subsection{Classification Accuracy}

We first compare against two prior FPGA-based implementations with on-chip learning capability: HaSiST \cite{b27} and TripleBrain \cite{b23} for their performance and hardware usage on the MNIST classification. Table \ref{tab:mnist} summarizes the results. 

Our on-chip learning achieves 95.92\% accuracy, surpassing TripleBrain's 95.10\%. This accuracy improvement can be attributed to our multilayer architecture with gradient-based learning, which enables richer feature representations than TripleBrain's single-layer STDP approach. While HaSiST achieves higher accuracy at 97.5\%, its training procedure operates on a surrogate ANN in full precision on a host computer, with trained weights subsequently transferred to the FPGA. Their in-hardware training engine can only handle $8\times8$ input. In contrast, our design performs both training and inference entirely on-chip in mixed precision, which is essential for edge scenarios requiring real-time adaptation without access to external computing resources.

\begin{table*}[t]
\centering
\caption{MNIST Accuracy Comparison Among FPGA-based On-chip Learning Implementations}
\label{tab:mnist}
\footnotesize
\begin{tabular}{lccccccccc}
\hline
\textbf{Impl.} & \textbf{Network} & \textbf{Synapses} & \textbf{Timesteps} & \textbf{FPGA} & \textbf{LUT} & \textbf{FF} & \textbf{DSP} & \textbf{BRAM} & \textbf{Acc.} \\
\hline
HaSiST$^*$ [27]     & 784-200-10   & 158,800  & 15  & Virtex-6    & 37.84 per synapse   & 2.63 per synapse   & -- & --  & 97.5\%  \\
TripleBrain$^\S$ [23]   & 784-256      & 200,704  & 200 & Zynq-7045   & 16,584$^\dagger$ & 33,168$^\dagger$ &32    & 131   & 95.10\% \\
Ours    & 784-100-100-10   & 89,400      & 10 & ZCU102      & 87,577   & 63,642 & 712   & 202  & 95.92\% \\
\hline
\multicolumn{10}{l}{\footnotesize $^*$HaSiST trains a surrogate ANN and transfers weights to SNN for inference. DSP and BRAM are not reported.} \\
\multicolumn{10}{l}{\footnotesize $^\S$TripleBrain use SOM-STDP and R-STDP. Resource reported as slices; TripleBrain uses a single-layer SNN with no multilayer gradient-based training.} \\
\multicolumn{10}{l}{\footnotesize $^\dagger$Converted from reported slices (Zynq-7000: 4 LUTs and 8 FFs per slice); upper-bound estimates.} \\
\end{tabular}
\end{table*}

For temporal datasets, we benchmark against the SNN models reported in [7] and [16]. The work in [7] trains SNN offline using BPTT and deploys inference on FPGA, while [16] implements SOLSA training on CPU/GPU. Both of them train the PSP kernel coefficients ($\alpha$ and $\beta$) in addition to the synapse weights ($w_{ij}$). To ensure a fair comparison, all models employ a three-layer structure ($D_{in}$ - 100 - 100 - $D_{out}$), where $D_{in}$ and $D_{out}$ denote the input and output dimensions of the dataset, respectively. As a reference, an LSTM model is also trained,  with its performance reported as a baseline representative of conventional artificial neural networks. Table \ref{tab:temporal} summarizes the results.


The results show a modest accuracy gap between our on-FPGA training and the SNNs trained on CPU/GPU. The difference can be attributed to the mixed-precision scheme and the adoption of a fixed PSP kernel. It also shows that the on-FPGA trained SNN performs comparable or even slightly better than the LSTM model. 

\begin{table}[t]
\centering
\caption{Temporal Dataset Accuracy Comparison}
\label{tab:temporal}
\begin{tabular}{lccccc}
\hline
\textbf{Impl.} & \textbf{RS} & \textbf{JV} & \textbf{BM} & \textbf{Epi} & \textbf{FM} \\
\hline
{[7]}$^\dagger$          & 0.869 & 0.969 & 1   & 0.961   & 0.62   \\
{[16]}$^\ddagger$        & 0.864 & 0.973 & 1   & 0.965   & 0.612   \\
LSTM                     & 0.877 & 0.935 & 0.925 & 0.839 & 0.625   \\
Ours                     & 0.842 & 0.959 & 1   & 0.956   & 0.61   \\
\hline
\multicolumn{6}{l}{\footnotesize RS=Racket Sports, JV=Japanese Vowels, BM=BasicMotions,} \\
\multicolumn{6}{l}{\footnotesize Epi=Epilepsy, FM=FingerMovements.} \\
\multicolumn{6}{l}{\footnotesize $^\dagger$Trained offline with BPTT; inference on FPGA.} \\
\multicolumn{6}{l}{\footnotesize $^\ddagger$Trained and evaluated on CPU/GPU.}
\end{tabular}
\end{table}

\begin{figure}[t]
  \centering
  \includegraphics[width=\linewidth]{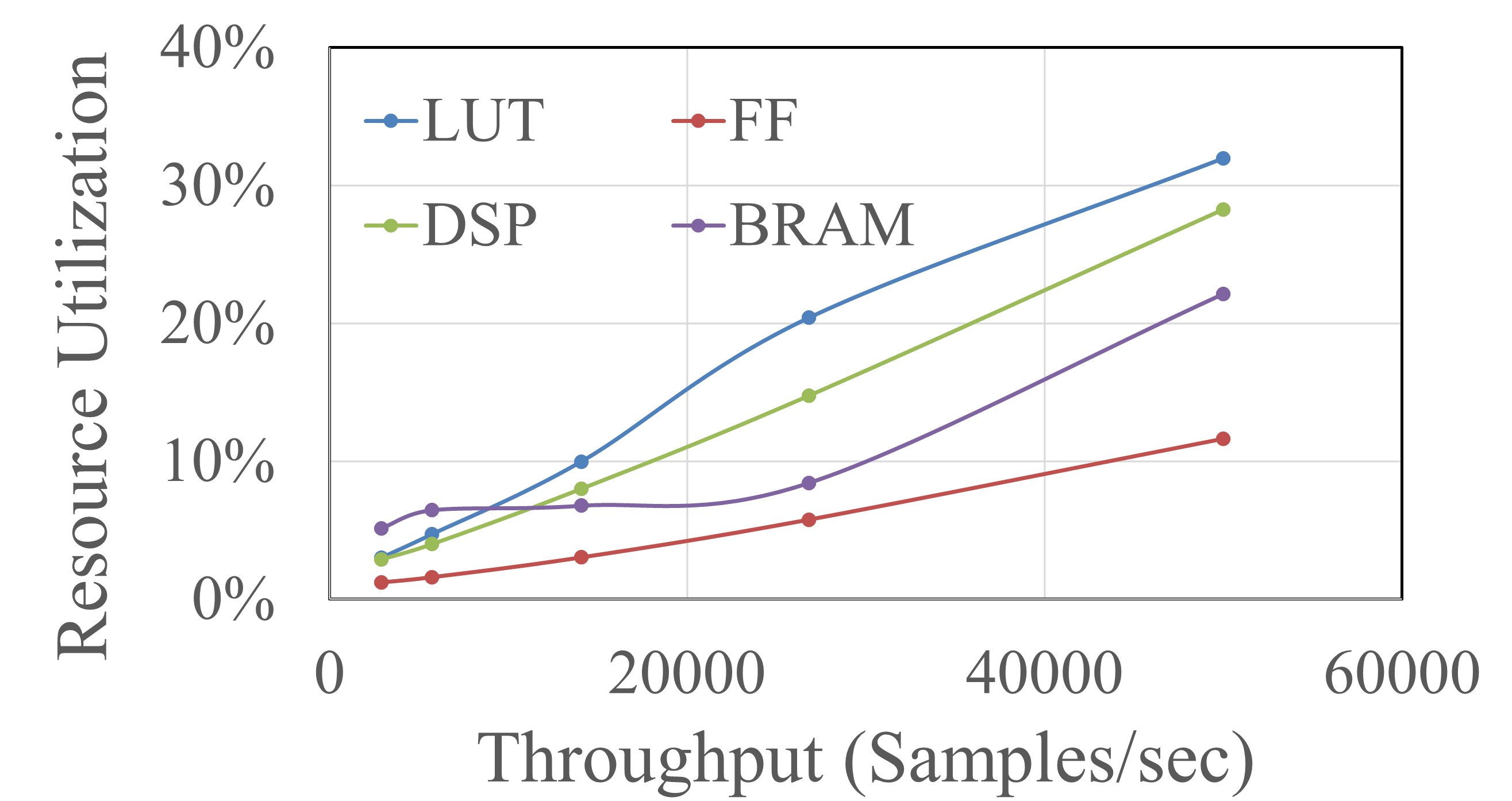}
  \caption{Hardware resource utilization(Racket Sports Dataset)}
  \label{fig:utilization}
\end{figure}

\subsection{Efficiency Comparison}
In this section, we evaluate the hardware efficiency, including throughput, clock frequency, and energy consumption, of our FPGA-based SNN against the baseline SNN model implemented on both CPU and GPU. The efficiency comparison is conducted on the temporal datasets, representative of typical sensor data rates in edge computing applications. The experimental setup included an Intel Core i7 8700 CPU and an NVIDIA RTX 2080 Ti GPU. To ensure a fair comparison, all selected platforms are manufactured using comparable semiconductor process nodes (the ZCU102 uses a 16nm FinFET process, the CPU uses 14nm, and the GPU uses 12nm), minimizing the influence of fabrication technology on the efficiency metrics.

As detailed in Table \ref{tab:efficiency}, our FPGA implementation achieves 2.5× to 6× throughput improvement for training while operating at a significantly lower clock frequency. Furthermore, it requires only 0.5~mJ per sample, an order of magnitude below the 5 and 8.125~mJ of the GPU and CPU baselines. Notably, the achieved throughput of 50,000 samples per second provides 5× to 50× headroom over the real-time sampling rates of the evaluated sensor datasets, confirming that our design can comfortably sustain online learning at real-time data rates in edge deployment scenarios.

\begin{table}[t]
\centering
\caption{Efficiency Comparison}
\label{tab:efficiency}
\begin{tabular}{lcccc}
\hline
\textbf{Platform} & \textbf{Samples/s} & \textbf{Clock (MHz)} & \textbf{Energy (mJ/sample)}  \\
\hline
FPGA (ZCU102)  & $\sim$50,000  & 125   & 0.5 \\
GPU (2080 Ti)  & $\sim$20,000  & 1,350 & 5  \\
CPU (Core i7)  & $\sim$8,000   & 3,600 & 8.125 \\
\hline
\multicolumn{5}{l}{\footnotesize Samples/s represents training throughput (forward + backward + update).}
\end{tabular}
\end{table}

\subsection{Hardware Resource Utilization}
In this section, we demonstrate the effectiveness of our loop-unrolling and memory management techniques and show that they produce Pareto optimum design points in resource and performance space. We varied the $C_{target}$ and $R_{max}$ to generate different loop unrolling strategies using Algorithm \ref{alg:unroll} to target distinct performance requirements. We then report the consumption of four primary resource types: Look-Up Tables (LUTs), Flip-Flops (FFs), Digital Signal Processing units (DSPs), and Block RAM (BRAM).

As illustrated in Fig. \ref{fig:utilization}, resource consumption scales linearly with throughput for most metrics, and nearly all data points lie on the Pareto front. The only exception is BRAM usage, which remains constant once throughput drops below a certain threshold, corresponding to a set of low unroll factors. In this regime, the number of BRAMs required is bounded by the amount of data that must be stored, a fixed value that cannot be further reduced through scaling. 
These results demonstrate the high flexibility of our architecture in terms of performance-hardware tradeoff. As a result, it can be seamless migrated to smaller, cost-effective FPGAs for lower-throughput tasks without requiring architectural modifications.

\section{Conclusion}
We presented an FPGA implementation of a multilayer SNN that performs both inference and online learning entirely on-chip. By combining a temporal forward-only training algorithm with a conservative FP32/FP16 mixed-precision scheme and a set of architecture-level optimizations---resource-aware loop unrolling, coarse-grained pipelining, and BRAM partitioning with hazard-resolving buffers---the design sustains gradient-based learning under tight edge resource budgets. On MNIST it reaches 95.92\% accuracy with fully on-chip training, surpassing TripleBrain (95.10\%) and approaching the off-chip-trained HaSiST (97.5\%), and stays within 3\% of BPTT-trained baselines on five temporal datasets. It delivers 2.5× to 6× the throughput of CPU and GPU platforms at roughly one-tenth the energy per sample, with headroom for real-time edge learning. The linear scaling of resource usage lets the same architecture migrate across FPGA capacities without modification.

\end{document}